\documentclass[conference]{IEEEtran}
\usepackage{graphicx}
\usepackage{float}
\usepackage{gensymb}
\usepackage[font=small,labelfont=bf]{caption}
\usepackage[utf8]{inputenc}
\usepackage[T1]{fontenc}
\ifCLASSINFOpdf
\else
\fi

\usepackage{eqparbox}
\usepackage{multirow}
\DeclareRobustCommand{\IEEEauthorrefmark}[1]{\smash{\textsuperscript{\footnotesize #1}}}

\begin{document}
%
\title{Pest Presence Prediction Using Interpretable Machine Learning}

\author{\IEEEauthorblockN{Ornela Nanushi$^*$\IEEEauthorrefmark{1}, Vasileios Sitokonstantinou$^*$\IEEEauthorrefmark{1,2}, Ilias Tsoumas\IEEEauthorrefmark{1} and Charalampos Kontoes\IEEEauthorrefmark{1}}

\IEEEauthorblockA{\IEEEauthorrefmark{1} National Observatory of Athens, IAASARS, BEYOND Center, Penteli, Greece\\
Email: \{ornela.nanushi, vsito, i.tsoumas, kontoes\}@noa.gr}

\IEEEauthorblockA{\IEEEauthorrefmark{2}Laboratory of Remote Sensing, National Technical University of Athens, Athens, Greece\\
}
}
\maketitle
\def\thefootnote{*}\footnotetext{These authors contributed equally to this work}

\begin{abstract}
Helicoverpa Armigera, or cotton bollworm, is a serious insect pest of cotton crops that threatens the yield and the quality of lint. The timely knowledge of the presence of the insects in the field is crucial for effective farm interventions. Meteo-climatic and vegetation conditions have been identified as key drivers of crop pest abundance. In this work, we applied an interpretable classifier, i.e., Explainable Boosting Machine, which uses earth observation vegetation indices, numerical weather predictions and insect trap catches to predict the onset of bollworm harmfulness in cotton fields in Greece. The glass-box nature of our approach provides significant insight on the main drivers of the model and the interactions among them. Model interpretability adds to the trustworthiness of our approach and therefore its potential for rapid uptake and context-based implementation in operational farm management scenarios. Our results are satisfactory and the importance of drivers, through our analysis on global and local explainability, is in accordance with the literature.  

\end{abstract}




%
\IEEEpeerreviewmaketitle

\section{Introduction}
Greece is the primary cotton producer for the European Union (EU) and the fifth-largest cotton exporter in the world; which makes cotton one of the most important crops for the national economy \cite{A1, tsiros2009assessment}. It is therefore imperative to protect cotton fields from pests and diseases with timely and effective actions to avoid damages. Helicoverpa armigera, also called cotton bollworm, is a serious threat to the crop, resulting in yield losses and suboptimal lint quality. This pest is widely abundant, especially in countries with warm or temperate climate like Greece \cite{A2}. In this work, we implement an interpretable machine learning approach to detect pest presence using satellite observations and meteorological data. By doing so, we identify the onset of harmfulness. In other words, consecutive presence estimations of the pest act as information about upcoming population peaks. This enables the farmers to optimize their pest management process.  

In order to successfully protect crops from pests, we need to understand their life cycle and thereby intervene before the abundance of insects becomes detrimental. Cotton bollworm's life cycle is relatively fast, and it takes place in four stages. First, the female moths lay the eggs. If the weather conditions are favourable, these eggs can hatch in less than three days. Then the eggs hatch and larvae emerge. This is the destructive phase of the pest. Larvae then turn into pupae, which are usually buried at a depth of $4$ to $10$ centimetres in soil and take $10$ to $15$ days to develop in a cocoon. An adult moth hatches out of the cocoon, ready to start the reproductive cycle all over again. Generally, it takes the pest approximately one month to complete its life cycle \cite{A3}.

Deep learning models, and mostly Convolutional Neural Networks (CNNs), have been widely used to recognize plant pests and diseases on plant images \cite{A12}. The aforementioned techniques work on close-up photos from the field that cannot be available frequently over large areas. As an alternative, one can use coarser spatial resolution data that can provide better coverage and temporal resolution, such as weather predictions and satellite images. For this, however, we need pest traps to know the occurrence and population of insects. There are different types of pest traps, i.e., pheromone, light or sticky traps etc. \cite{B2}. Using such traps renders the problem quantitative, i.e., the interest no longer lies on the pest or damage recognition, but on the estimation of its population.

Several studies have worked on estimating pest population (regression) \cite{A9} or detecting significant presence (binary classification) \cite{A7,A8}. With reference to the latter, knowing the exact population of pests is not always crucial. Instead, action thresholds are used to classify the amount of pests as harmful or not \cite{A5}. One can find approaches that make use of i) physical models \cite{A10} or ii) data driven models that employ remote sensing and in-situ measurements \cite{A7,A11}. It is worth noting that most related studies use weather data, whereas in fewer cases remote Earth observations are used to capture the changes caused by the pests, but also the favourable vegetation conditions for their occurrence \cite{A11}.  Recurrent Neural Networks (RNN) have been applied to weather data time-series, accounting for the temporal evolution of features and capturing the cyclical nature of pest abundance \cite{A4}. In other words, when trying to predict pest occurrence at a given time instance, one cannot ignore the weather conditions and the vegetation status of previous days \cite{A13}.  

Traditional machine learning is also widely used in the literature, with particular focus on regression models. In \cite{A20}, the authors perform multivariate regression analyses, and they find that temperature and  rainfall have significant correlations with the cotton bollworm population. Similarly, in \cite{A14} they find temperature, wind speed and sunshine hours to be important. Furthermore, in \cite{A6} they perform binary classification for pest occurrence detection using relative humidity and temperature. Complementary to weather factors, some studies also use host plant phenology represented by the Normalized Difference Vegetation Index (NDVI) or other vegetation indices \cite{A7}. This is done in order to study the association between the crop's growth and the pest population. Satellite derived vegetation indices are also used in \cite{A16}, where they introduce an ecology oriented model for population dynamics. 

Motivated by real life requirements, we designed and implemented a predictive method to detect harmful cotton bollworm presence in cotton fields. Our model is interpretable, which means that one can understand how it arrived to a specific decision and which are the drivers that mattered the most. This allows the user of the model's outputs (e.g., farmers) to understand and trust them and therefore adhere to the recommendations, and even fine-tune them with their expert knowledge. Our approach makes use of both vegetation indices and weather data, and to the best of our knowledge there is no similar work for cotton bollworm presence estimation. We perform two experiments, i.e., one that makes use of past trap catches and one that does not, yielding satisfactory results for both. The latter experiment is of great significance as it allows for the method to be applicable anywhere in space, irrespective of the presence of pest traps in the vicinity. 

\section{Methodology}
\begin{figure}[!h]
  \includegraphics[scale=0.419]{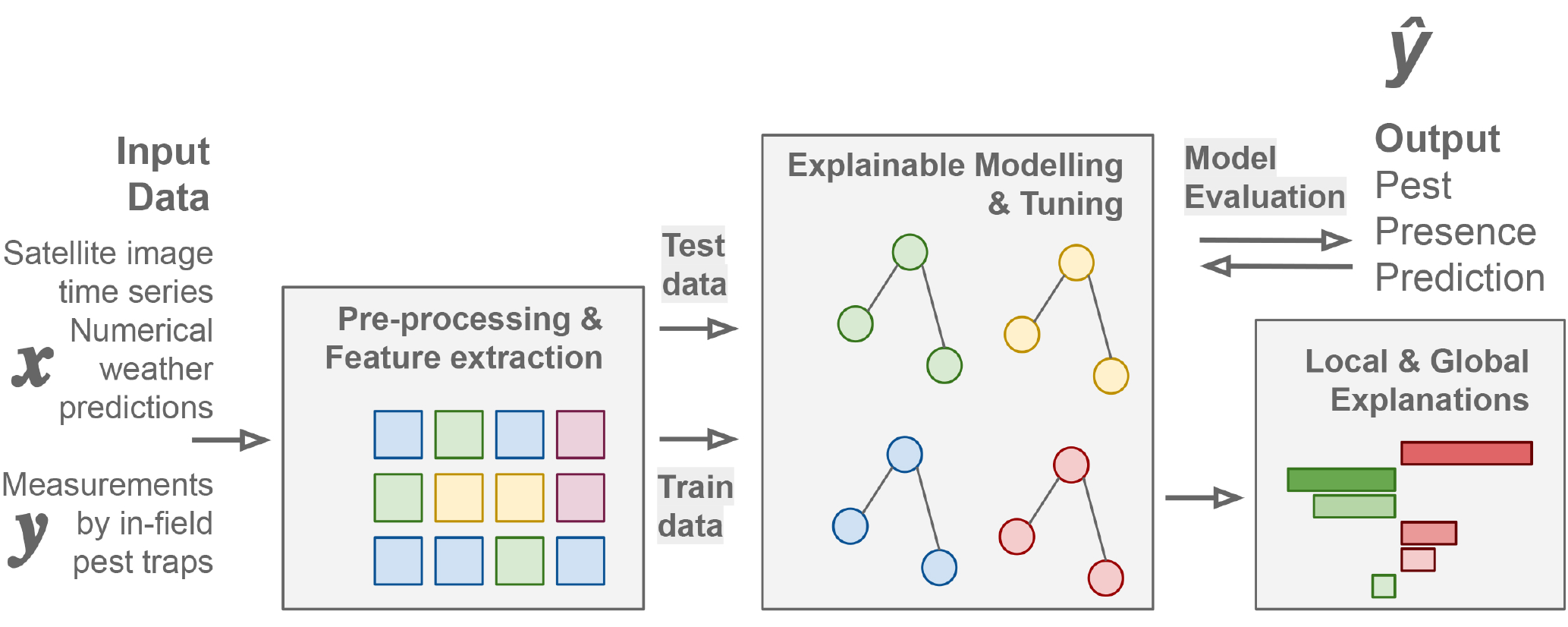}
  \caption{Workflow of the method for bollworm presence prediction. \label{fig:firstfigure}}
  \vspace{-2em}
\end{figure}

Our work focuses on the prediction of cotton bollworm presence based on meteorological data, satellite earth observations and past trap catches. This section elaborates on the formulation of the problem, and the acquisition, pre-processing and engineering of the data to analysis-ready features. Fig. \ref{fig:firstfigure} shows an overview of the steps followed. 
\vspace{-1em}

\subsection{Problem formulation}

Let $T$ be the set of traps we have at our disposal. We note as $t_{ij}\in T$ the entity describing trap $i$ on date $j$. Every $t_{ij}$ can be expressed as
$t_{ij} = (x_{ij},y_{ij})$, where $x_{ij}\in{\rm I\!R^{d}}$ is a $d$ dimensional vector containing the corresponding weather and satellite features, and $y_{ij}\in\{0, 1\}$, with $0$ referring to pest absence and $1$ referring to pest presence. Our aim is to find a model $M:{\rm I\!R^{d}} \rightarrow \{0, 1\} $, so that for any trap $t_{ij}$, to predict the onset of harmful presence of bollworm population that will in turn signal the time for intervention. 


\textbf{Explainable classifier approach:} There exist many sophisticated models that could be appropriate for our problem. However, we should bear in mind that in the agricultural sector, whoever be the user, it is common to have substantial empirical knowledge. For example, someone who cultivates cotton is likely to know when to sow, the crop's enemies, the favourable conditions etc. Therefore, it is crucial to make these users trust the model and its outputs \cite{inproceedings}. Furthermore, if the users have some insight on the main drivers and find the result unjustifiable, they can disregard the prediction or frame it in context and then take appropriate action. The above examples point out the need for model transparency, which can be achieved using explainable machine learning \cite{A17}. We therefore chose to proceed using an interpretable classifier. 


The model we used is the Explainable Boosting Machine (EBM), as implemented in the InterpretML framework. In machine learning, there is often a trade-off between accuracy and intelligibility. This is not true for EBM that achieves a performance comparable to powerful black-box models, while offering global and local explanations on the predictions \cite{unknown2}. EBM is an augmented version of the Generalized Additive Models (GAMs), which are expressed by (\ref{eq1}).


\begin{equation}
\label{eq1}
g(E(y)) = \beta_{0} + \sum{f_j(x_j)}
\end{equation}
In practice, EBM is a GAM model with interactions based on the GA2M method, expressed by (\ref{eq:2}) \cite{pairwise}.

\begin{equation} \label{eq:2}
g(E(y)) = \beta_{0} + \sum{f_j(x_j)} + \sum{f_{ij}(x_i,x_j)}  
\end{equation}

Every feature function $f_{j}$ is learnt using many shallow decision trees. The learning is achieved using one predictive feature at a time in round-robin pass over the train data, while performing gradient updates. In more detail, we train a tree on the first feature, then in boosting fashion we update the residual and move to the second feature, then we train another tree and so on. The learning rate is low, and thus the feature order does not matter. Interaction functions, $f_{ij}$, are learnt via the FAST method \cite{pairwise}.

All features' contributions, together with the interactions, yield a logit score. Scores are then summed and passed to the link function $g$, which is chosen depending on the context. In binary classification, as in our case, EBM uses the logarithmic loss. Since the model is additive, the contributions of the terms can be sorted and visualized. This way, we can understand the main drivers of the model and produce new domain knowledge, but also enhance the trust on the outputs. 


\subsection{Data collection}

\textbf{Trap data:} To obtain insect population measurements, we collaborated with the company Corteva Agriscience Hellas that has a trap network for helicoverpa armigera. The network consists of pheromone traps in 26 different locations in the wider region of Central Macedonia, Greece. The traps are located in such a distance so as to not interact with each other and are examined (trapped insect counting) by a specialist every 3-5 days. The observations take place from June 2020 until early September 2020 and June 2021 until early September 2021. The trap data include i) the trap locations, ii) the number of pests on each trap and iii) the Day of Year (DoY) they were recorded. In total, there are 10 locations for 2021 and 16 locations for 2020. 

\textbf{Meteorological data.}
In real life scenarios, exact weather conditions are very hard to obtain. This is true due to the absence of a dense network of weather stations. To overcome this, we use numerical weather predictions using our own configuration of WRF-ARW \cite{A18, sitokonstantinou2021semi}. The model is of high spatial resolution (2 km), and predictions are made hourly. For each trap location, we obtained daily values for air (2 m) and soil temperature (0 m), relative humidity (RH), accumulated precipitation (AP), dew point (DP), and wind speed (WS). These parameters have been extensively used in related work. 

\textbf{Satellite data.} 
We used Sentinel-2 images to capture the vegetation status (at the location of the traps) and its evolution through time \cite{A21}. Using atmospherically corrected Sentinel-2 images, we calculated a number of vegetation indices that highlight particular characteristics of the crops (e.g., moisture, physiological stress etc.) that are known to i) drive the occurrence of pests, but also ii) capture the vegetation cover changes caused by the insects \cite{VI}. In detail, we used the normalized difference vegetation (NDVI), water (NDWI), moisture (NDMI), greenness (GI) and greenness chlorophyll (GCVI) indices. 



%


\subsection{Feature engineering} \label{features}
We provide pest presence predictions for each trap and for each date that we have catches. Each of those instances in time and space (trap/inspection day, 526 instances) is described by a series of features that are engineered using the data variables described earlier. Regarding the trap data, we use the coordinates of the trap locations, the corresponding dates they were visited and the number of insects observed per visit. In the feature space, we included the catches of the three latest visits prior to the day of prediction. The day of visit was encoded using the sine and cosine of the DoY, which is common for cyclical features. The number of catches for the day of prediction function as the labels for training and evaluation. The labels are binarized to pest presence or absence according to a certain threshold ($t=10$), which represents the number of insects above which the pest is considered harmful to the crop, also known as the action threshold. 



The meteorological and vegetation index variables were engineered into accumulated features to capture the near-past (7 days) information that drives pest occurrence. Specifically, the accumulated vegetation indices were calculated using the cumulative integral of the time-series curve, whereas the accumulated weather features were calculated by summing the daily values over the last 7 days. Growing degree days, which capture the effective growth time of the plant,  were also calculated according to (\ref{eq:3}) \cite{A19}. $T_{base}$ refers to the temperature under which the cotton does not develop and is equal to 15.6$^{\circ}$
C, and $T_{max}$, $T_{min}$ are the maximum {\&} minimum air temperatures (2 m), respectively.
\begin{equation} \label{eq:3} GDD = \max\left({\frac{T_{max}+T_{min}}{2}-T_{base},0}\right) \end{equation}
The data was standardized using a standard scaler. The EBM was run using 100 inner bags, 100 outer bags, and a learning rate of 0.01 that according to \cite{unknown2} is an appropriate tuning to achieve both high accuracy and interpretability. 

\section{Experiments {\&} Results} 

As described in Sec. \ref{features}, we used cumulative integrals of vegetation indices, accumulated and current weather data, and past trap catches to predict the occurrence of bollworms. We also ran experiments without including the insect catches in order to explore the capability of our model to predict pest presence anywhere in space, and not just for regions for which we have traps. Table \ref{tab:ms} depicts the accuracy, AUC and F1 score for both cases, averaged over 10 random train/test splits (70/30).
\vspace{-1em}
\begin{table}[!ht]
\caption{Performance of the pest prediction model for Case A with all features, and Case B without trap related features.}
\label{tab:ms}
\centering
\scalebox{1.3}{
\begin{tabular}{c|c|c|c|}
\cline{2-4}         & \textbf{Class} & \textbf{Case A} & \textbf{Case B} \\ \hline
\multicolumn{1}{|c|}{\textbf{Accuracy}} & -              & 0.75 ± 0.03     & 0.69 ± 0.02     \\ \hline
\multicolumn{1}{|c|}{\textbf{AUC}} & -              & 0.83 ± 0.03     &  0.77 ± 0.03     \\ \hline
\multicolumn{1}{|c|}{\multirow{2}{*}{\textbf{F1 score}}} & Pest absence & 0.75 ± 0.04 & 0.68 ± 0.03 \\ \cline{2-4} 
\multicolumn{1}{|c|}{}                  & Pest presence  & 0.74 ± 0.04     & 0.69 ± 0.02     \\ \hline

\end{tabular}}
\end{table}
\vspace{-1.4em}


As expected, Case A considerably outperforms Case B. It should be noted, however, that Case B showcases promising performance given the model was able to predict pest presence using only earth observations and weather data. In the context of an operational pest management service, the Case B approach has great value, as it can be applicable to any cotton field without the need of being near a pest trap. 

\begin{figure}[h!]
\centering
\includegraphics[scale=0.39]{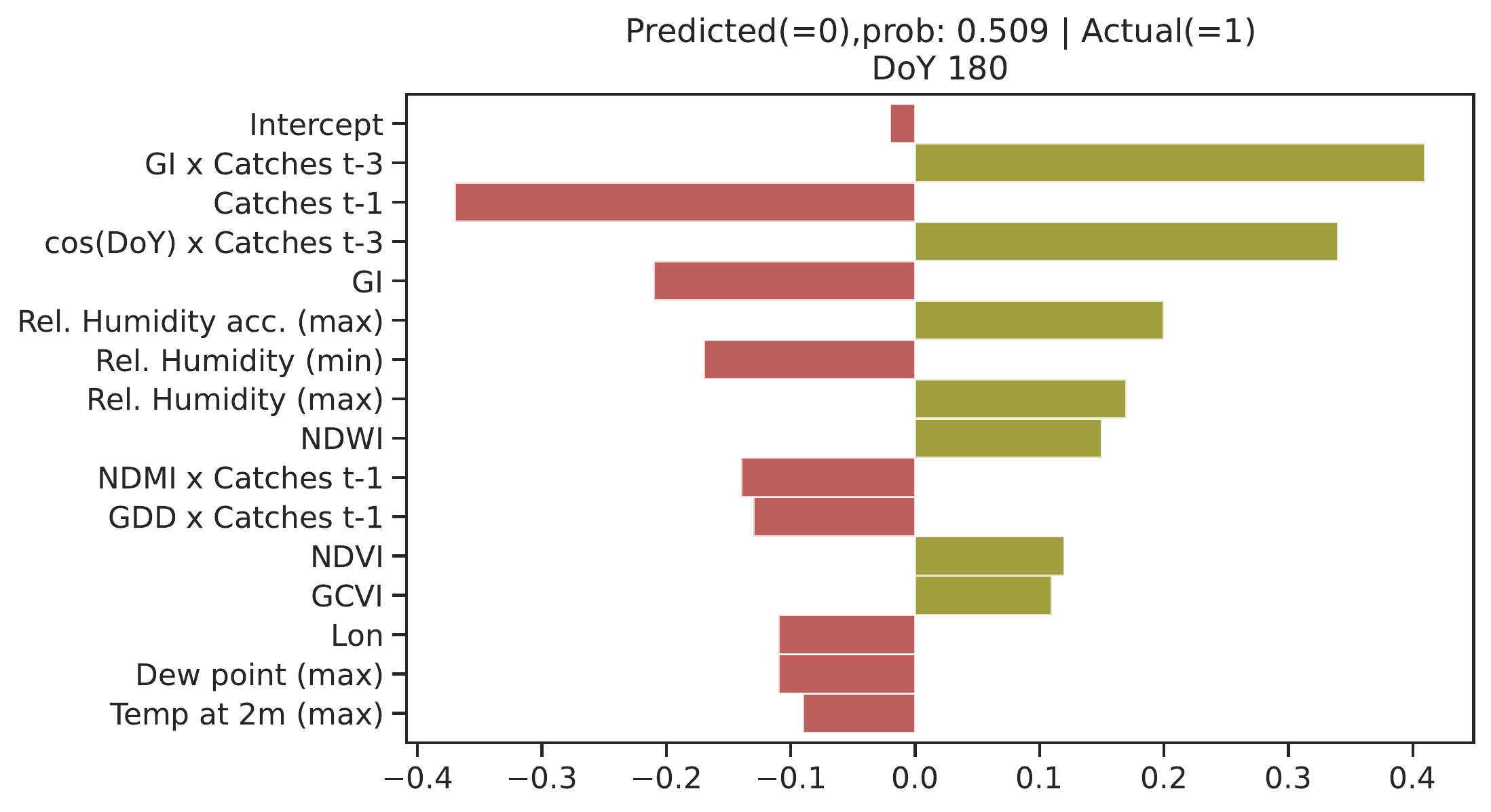} 
\caption{Local explainability plot for an uncertain prediction.}\label{fig:fourthfigure}
\vspace{-1em}
\end{figure}

The pest presence threshold, set at 10 catches, amounts to approximately 3 insects a day, as traps are visited every 3-5 days. According to local agronomists that we consulted, a couple of catches a day are not considered harmful. Nevertheless, the threshold is only an approximation. Fig. \ref{fig:fourthfigure} illustrates the seriousness of the errors of our model (Case A) by visualizing how close to the action threshold they occurred. The left histogram depicts the distribution of catches for the model errors. As expected, most errors are situated near the action threshold. 

\begin{figure}[h!]
\centering
\includegraphics[scale=0.435]{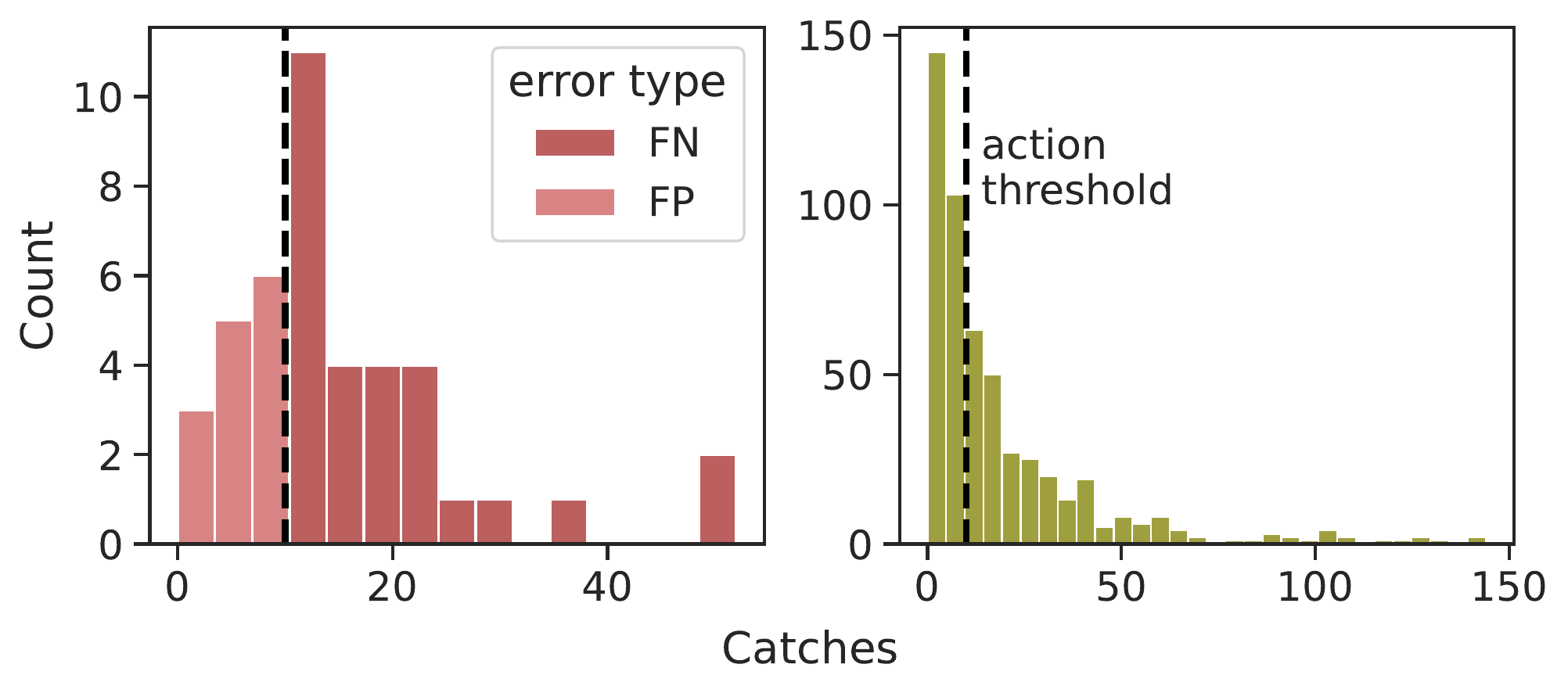} 
\caption{Distribution of the number of catches for all trap measurements (right) and for the model errors (left).\label{fig:secondfigure}}
\vspace{-1.4em}
\end{figure}

\begin{figure}[h!]
\centering
\includegraphics[scale=0.51]{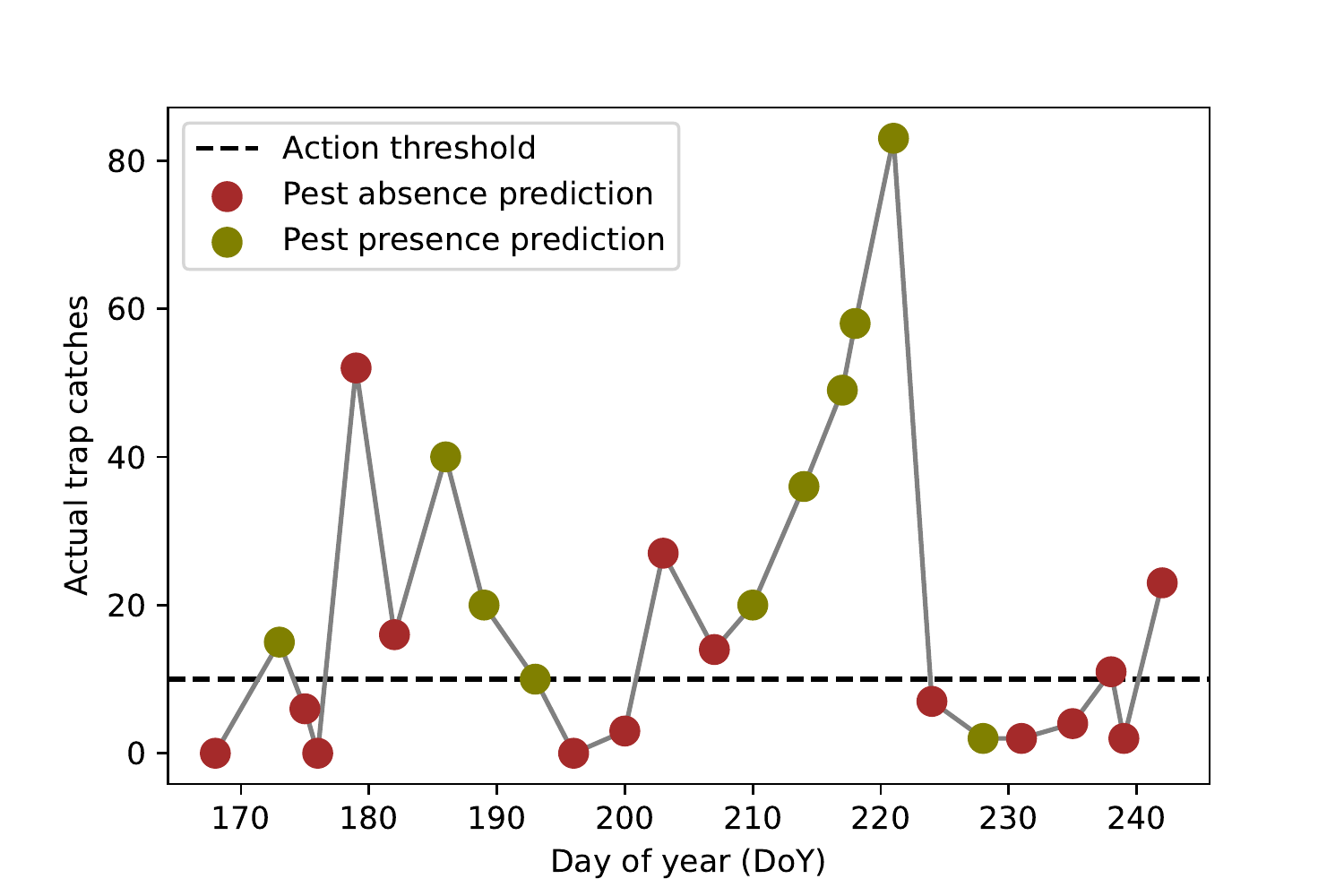} 
\caption{An indicative time-series of insect catches from a single trap in Macedonia, Greece. The green dots are the pest presence predictions and the red dots are the pest absence predictions. The dashed line represents the action threshold set at $t=10$. }\label{fig:thirdfigure}
\vspace{-1.4em}
\end{figure}

Using the InterpretML framework, we extracted local and global explainability reports. Global explanations account for all model predictions, whereas local explanations are provided for each individual prediction. Table \ref{tab:global_exp_table} shows the global explanations for Cases A and B. The scores indicate how much each feature affected the result. For Case A, the features with the strongest impact were the past trap catches. This is expected since pest presence depends on the population of the previous days. For Case B, it is mostly the weather features, both current and accumulated, as well as their interactions (acting as a new parameter) that compensate for the lack of trap information. Wind speed, relative humidity, precipitation, air, and soil temperature appear to be very important features for both cases. This is also supported by the literature \cite{A20,A14,A16}. Time features, such as cos(DoY) and sin(DoY), are also important since there is clear seasonality in bollworm population (see Fig. \ref{fig:thirdfigure}). Finally, the vegetation indices, and interactions of them with weather features for Case B, are also found among best features. This supports the choice of combining satellite earth observations with weather data, which has not been done before for bollworm presence estimation.

\vspace{-0.9em}
\begin{table}[!ht]
\caption{Feature importance according to EBM's global explainability report for Case A with all features and Case B without trap related data.}
\label{tab:global_exp_table}
\centering
\scalebox{1.1}{
\begin{tabular}{|cc|cc|}
\hline
\multicolumn{2}{|c|}{\textbf{Case A}}      & \multicolumn{2}{c|}{\textbf{Case B}}                     \\ \hline
\multicolumn{1}{|c|}{\textbf{Features}} & \textbf{Scores} & \multicolumn{1}{c|}{\textbf{Features}}              & \textbf{Scores} \\ \hline
\multicolumn{1}{|c|}{Catches t-1}   & 0.29 & \multicolumn{1}{c|}{WS (min) - AP acc.}           & 0.13 \\ \hline
\multicolumn{1}{|c|}{Catches t-2}       & 0.16            & \multicolumn{1}{c|}{WS (max) - T at 0 m (max) acc.} & 0.12            \\ \hline
\multicolumn{1}{|c|}{T at 2 m (min)}    & 0.15            & \multicolumn{1}{c|}{RH (min) acc. - GDD}            & 0.12            \\ \hline
\multicolumn{1}{|c|}{cos(DoY)}      & 0.13 & \multicolumn{1}{c|}{cos(DoY)}                     & 0.11 \\ \hline
\multicolumn{1}{|c|}{GI}            & 0.13 & \multicolumn{1}{c|}{T at 0 m (max) acc. - cos(DoY)} & 0.11 \\ \hline
\multicolumn{1}{|c|}{RH (max)}      & 0.13 & \multicolumn{1}{c|}{WS (max)}                     & 0.10 \\ \hline
\multicolumn{1}{|c|}{RH acc. (max)} & 0.12 & \multicolumn{1}{c|}{NDMI - WS (min)}              & 0.09 \\ \hline
\multicolumn{1}{|c|}{NDWI}          & 0.11 & \multicolumn{1}{c|}{GCVI}                         & 0.09 \\ \hline
\multicolumn{1}{|c|}{Lon}           & 0.11 & \multicolumn{1}{c|}{NDVI - NDMI}                  & 0.09 \\ \hline
\multicolumn{1}{|c|}{GCVI}          & 0.10 & \multicolumn{1}{c|}{GCVI - sin(DoY)}              & 0.09 \\ \hline
\end{tabular}}
\end{table}

The aforementioned experiments, used train and test data randomly selected from all the available traps. Nevertheless, it is also important to observe the temporal pattern of our predictions. The idea is to identify when the presence is significant, indicating the population peak over the next days and thus the need to intervene. To do so, we needed to see how the model would perform when tested on the entire time-series of a trap. Specifically, instead of training our model (Case A) using samples from  all the traps, we would leave out one trap at a time and use it as a test set. Fig.\ref{fig:thirdfigure} shows the actual trap catches against time, together with the action threshold that indicates whether a point should have been classified as pest presence or not. This is an indicative example for a single trap located in Central Macedonia, Greece. The green dots represent pest presence predictions, the red dots represent pest absence predictions, and the dashed line represents the action threshold ($t=10$). We can conclude that the model performs rather well for this particular trap. The principal troughs and peaks are clearly identified, indicating that the model is able to predict the onset and termination of pest harmfulness. It is also worth noting how the model was able to identify the end of the bollworm population cycle after DoY 220. 
\vspace{-1em}

To further investigate this test trap, we used the local explanations of the InterpretML framework. We indicatively examine the prediction for DoY=180. This one, despite being a peak, was not detected as such. Looking at the local explainability graph in Fig. \ref{fig:fourthfigure}, we can gain insight on this decision. The coloured bars indicate which features drove the decision towards pest presence (green) and which features towards pest absence (red). The predicted class is assigned with a probability of only 0.509, indicating uncertainty. The feature which had the strongest impact towards the true label was the interaction of GI with Catches t-3. This is very logical since the catches at the instance t-3 indicated pest presence. On the other hand, the catches at the instance t-1 strongly suggest pest absence. By examining Fig. \ref{fig:thirdfigure}, we can see that the pest catches in the previous visit were zero and hence unlikely to have strong pest presence only four days later.

\section{Conclusions and future work}

In this work, we used an interpretable classifier to detect significant cotton bollworm presence to indicate the onset of pest harmfulness. This is the first study that uses both weather data and earth observations to model the occurrence of this pest. We ran two experiments: i) one that incorporates past trap catches in the feature space and ii) one that uses only meteorological parameters and vegetation indices. Both approaches yielded satisfactory results, with model (i) outperforming model (ii). Model (ii), however, is of particular interest in the context of pest management, as it can be applicable anywhere in space without being dependent on the presence of pheromone (or other) traps. Moreover, by using the InterpretML framework, we can understand and trust the model's decisions. Not only we predict the onset of the rise in pest population, but we also provide significant insight on the main drivers of the model's outcomes. This makes our approach transparent and easy to understand. 

Future work includes forecasting pest presence for the next two to five days using numerical weather predictions and projections of vegetation indices. Moreover, using a larger trap dataset we can enhance the performance of model (ii) and thus expand inference spatially, in areas where no traps are set, resulting in pest presence-absence maps.


\section*{Acknowledgments}
This work has been supported by the EU H2020 projects e-shape (No 820852) {\&} EIFFEL (No 101003518). The authors would like to acknowledge Corteva Agriscience Hellas S.A. for providing the trap data and Nikolaos S. Bartsotas for providing the numerical weather predictions.



\bibliographystyle{IEEEtran}
\bibliography{refs}
%
%
%

\end{document}